\documentclass{article}

\usepackage{lineno}

\usepackage[final]{nips_2018}

\usepackage{graphicx}  

\usepackage{url}  

\usepackage[dvipsnames]{xcolor}

\usepackage{subcaption}
\usepackage[lined, boxed, ruled, commentsnumbered, noend]{algorithm2e}

\usepackage{amssymb,amsmath,amsthm}
\usepackage{mathtools}
\usepackage{xspace}

\usepackage[makeroom]{cancel}

\SetCommentSty{mycommfont}

\newcommand{\denselist}{
  \itemsep 0pt \topsep-8pt\partopsep-8pt
}

\DeclareMathOperator*{\argmax}{arg\,max}

\theoremstyle{definition}

\theoremstyle{remark}

\numberwithin{equation}{section}

\newcommand{\figref}[1]{Fig.~\ref{#1}}

\newcommand{\secref}[1]{\S\ref{#1}}

\newcommand{\lnref}[1]{Line~\ref{#1}}
\newcommand{\algref}[1]{Algorithm~\ref{#1}}

\newcommand{\algname}{\textsf{MF-MI-Greedy}\xspace}
\newcommand{\sfgpopt}{\textsf{SF-GP-OPT}\xspace}
\newcommand{\explorelf}{\textsf{Explore-LF}\xspace}

\newcommand{\sfgpucb}{\textsf{GP-UCB}\xspace}
\newcommand{\sfmves}{\textsf{MVES}\xspace}
\newcommand{\sfest}{\textsf{EST}\xspace}

\newcommand{\sfgpmi}{\textsf{GP-MI}\xspace}

\newcommand{\nan}{\textsf{null}\xspace}

\newcommand{\mean}{\mu}
\newcommand{\std}{\sigma}
\newcommand{\cov}{k}

\usepackage{dsfont}

\newcommand{\expctover}[2]{\mathbb{E}_{#1}\!\left[#2\right]}

\newcommand{\cS}{{\mathcal{S}}}

\newcommand{\cL}{{\mathcal{L}}}

\newcommand{\by}{{\mathbf{y}}}

\newcommand{\paren} [1] {\ensuremath{ \left( {#1} \right) }}

\newcommand{\bigOmega}[1]{\ensuremath{\Omega\paren{#1}}}

\newcommand{\reals}{\ensuremath{\mathbb{R}}}

\newcommand{\GP}[1]{\text{GP}\paren{#1}} 

\newcommand{\reward}{q} 

\newcommand{\fidelity}{f} 
\newcommand{\utility}{f} 

\newcommand{\exDom}{\mathcal{X}} 

\newcommand{\ex}{x} 
\newcommand{\obs}{y} 
\newcommand{\bobs}{\by} 

\newcommand{\selected}{\cS} 
\newcommand{\eplow}{\cL} 

\newcommand{\noise}{\varepsilon}

\newcommand{\action}[1]{\langle #1\rangle} 

\newcommand{\fid}{\ell}
\newcommand{\targetfid}{m} 

\newcommand{\normal}{\mathcal{N}}

\newcommand{\tarf}{f_\targetfid} 

\newcommand{\budget}{\Lambda} 
\newcommand{\Cost}{\Lambda} 

\newcommand{\costof}[1]{\lambda_{#1}} 

\DeclarePairedDelimiterX\parencond[2]{(}{)}{#1 \;\delimsize\vert\; #2}
\DeclarePairedDelimiterX\bracketcond[2]{\{}{\}}{#1 \;\delimsize\vert\; #2}

\newcommand{\condinfgain}[2]{\mathbb{I}\parencond*{#1; \tarf}{#2}} 
\newcommand{\condentropy}[2]{\mathbb{H}\parencond*{#1}{#2}} 

\newcommand{\policy}{\pi} 

\title{Optimizing Photonic Nanostructures via Multi-fidelity Gaussian Processes}

\author{
  \textbf{Jialin Song}\textsuperscript{\textdagger} \qquad
  \textbf{Yury S. Tokpanov}\textsuperscript{\textdagger} \qquad
  \textbf{Yuxin Chen}\textsuperscript{\textdagger} \qquad
  \textbf{Dagny Fleischman}\textsuperscript{\textdagger} \\
  \textbf{Kate T. Fountaine}\textsuperscript{\textdagger\textdaggerdbl} \qquad
  \textbf{Harry A. Atwater}\textsuperscript{\textdagger} \qquad
  \textbf{Yisong Yue}\textsuperscript{\textdagger} \\
  \\
  \textsuperscript{\textdagger}Division of Engineering and Applied Sciences \\
  California Institute of Technology
  \\
  \texttt{\{jssong, tokpanov, chenyux, haa, yyue\}@caltech.edu}\\ \texttt{dagny.fleischman@gmail.com}\\
  \\
  \textsuperscript{\textdaggerdbl}NG Next \\
  Northrop Grumman Corporation\\
  \texttt{katherine.fountaine@ngc.com}\\
}

\begin{document}

\maketitle

\begin{abstract}
 We apply numerical methods in combination with finite-difference-time-domain (FDTD) simulations to optimize transmission properties of plasmonic mirror color filters using a multi-objective figure of merit over a five-dimensional parameter space by utilizing novel multi-fidelity Gaussian processes approach. We compare these results with conventional derivative-free global search algorithms, such as (single-fidelity) Gaussian Processes optimization scheme, and Particle Swarm Optimization---a commonly used method in nanophotonics community, which is implemented in Lumerical commercial photonics software.
 We demonstrate the performance of various numerical optimization approaches on several pre-collected real-world datasets and show that by properly trading off expensive information sources with cheap simulations, one can more effectively optimize the transmission properties with a fixed budget.
\end{abstract}


\section{Introduction}

Designing compact integrated color filters with ultra-narrow bandwidth is of great importance for realizing practical multispectral and hyperspectral imaging. Each pixel of a hyperspectral imaging device records the spectrum of light from the environment, providing significantly more information in comparison with conventional imaging techniques, that can help with, e.g.,  materials identification or objects detection. Such devices can find a wide range of applications in different areas of science and technology, including medicine, material science, astronomy, environment monitoring to name a few.

Surface plasmon polaritons (SPPs) allow for extreme miniaturization of integrated photonic devices via strong light confinement that can result in very small wavelength of light (potentially dozens of times smaller than in free space) \citep{atwater2001plasmonics}. Hence, such plasmonic devices look very promising as a platform for designing ultra-compact integrated narrow-band photonic filters \citep{yokogawa2013filter,chen2012refl}. Periodic arrays of subwavelength holes or nanoslits in metal films enable efficient excitation of SPPs by satisfying momentum-matching with the addition of a grating wavevector. The grating materials, geometry, and symmetry control the excitation efficiency \citep{ebbesen1998eot}.  In particular, periodic arrays of subwavelength apertures passing through an optically thick metal film exhibit enhanced transmission exclusively at conditions corresponding to constructive mutual interference between incident light and SPPs traveling along the surface between adjacent slits and acting as a band-pass color filter \citep{dagny2017filter}. However, periodicity of slits can be effectively achieved by placing reflective mirrors around a slit.

Designing such plasmonic mirror filters sets up a non-trivial optimization challenge: number of independent parameters can easily exceed a few dozens, optimization landscape itself is non-convex with many local minima. Additional challenge arises from the fact that the derivation of analytical model is nearly impossible due to near-field effects and complicated geometry, hence numerical simulations of underlying physical processes (governed by Maxwell's equations) are required.

Here we present an application of an in-house developed optimization strategy based on multi-fidelity Gaussian processes \cite{song2018general} to this nanophotonics design problem. Our simulation setup allows us to easily control the fidelity of numerical calculations by changing geometric mesh size and total time duration of simulated physical processes. We compare it with a conventional Gaussian Processes approach and a commonly used algorithm---Particle Swarm Optimization, which is implemented in Lumerical commercial nanophotonics software\footnote{https://kb.lumerical.com}.




\section{Background and Related Work}

\subsection{Background on Nanophotonics Optimization Problem}


We employ numerical methods to solve for an optimized design of this plasmonic color filter.  We use Lumerical commercial implementation of the finite-difference time-domain (FDTD) method to simulate the transmission spectra of such devices. In post-processing analysis, we extract observable scalar figures of merit corresponding to the goodness of the spectrum such as transmission peak amplitude, its offset from designed wavelength, full-width half-maximum, signal-to-noise ratio, and combine them into one weighted figure of merit (FOM). Using this FOM, the evolution of the design can then be pursued as a minimization problem over a geometric parameter space, which can be driven using any of a variety of iteration schemes. Here, we will search over a five parameter domain, describing the geometry of our devices.

The filters center wavelength, linewidth, and amplitude are determined by the interaction of several physical processes including the amplitude and phase of photon-plasmon coupling at the slits, the strength of mutual coupling between the propagating waveguide channels, and the effective index of each participating electromagnetic mode.  Due to the interplay of these several physical resonances, the FOM corresponding to our filters exhibits oscillations in parameter space that are likely to trap a local directed search method in a globally non-optimal local solution. Therefore, gradient descent or other local optimization modalities are excluded for this purpose. Conversely, the relatively large computational cost of the FDTD forward problem limits the applicability of purely stochastic approaches like evolutionary methods. Instead, we seek methods which execute global, derivative-free search with an efficient iteration strategy that calls the forward problem solver a limited number of times (under defined budget).

\subsection{Numerical Approaches to Black-box Optimization}
In this section, we review related numerical approaches for black-box optimization. 

\subsubsection{Classical Approaches on Single-fidelity Blackbox  Optimization}

\paragraph{Heuristics} 
There are many stochastic heuristics for finding approximate solutions of non-convex optimization problems, such as simulated annealing \citep{kirkpatrick1983annealing}, genetic algorithms \citep{holland1984genetic}, particle swarm optimization (PSO) \citep{kennedy1995pso} and many others. In this paper, we are using PSO as one of the baselines for comparison, as it has a wide use in nanophotonics community \citep{pso2008mutitu, shokooh2010leaky} and it is implemented in Lumerical commercial nanophotonics software.

In the particle swarm algorithm, the potential solutions, called particles, are initialized at random positions and velocities, and then move within the parameter search space. 
The particles are subject to three forces as they move: spring force towards the personal best position ever achieved by that individual particle, spring force towards the global best position ever achieved by any particle, a frictional force, proportional to the velocity. At each iteration velocities are stochastically updated based on these forces and previous velocity values, then new particles locations are calculated as the old ones plus the velocities, modified to keep particles within bounds. The algorithm proceeds until a specified stopping criterion is met.

PSO is inspired by the behavior of animal aggregations like flocks of birds or insects swarming. Each particle is attracted to some degree to the best location it has found so far, and also to the best location any member of the swarm has found. After some steps, the population can coalesce around one location, or can coalesce around a few locations, or can continue to move.

\paragraph{Gaussian processes optimization}
Optimizing an unknown and noisy function is a common task in Bayesian optimization. In real applications, such functions tend to be expensive to evaluate, for example tuning hyperparameters for deep learning models \citep{snoek2012practical}, so the number of queries should be minimized.  As a way to model the unknown function, Gaussian process (GP) \citep{rasmussen:williams:2006} is an expressive and flexible tool to model a large class of functions. A classical method for Bayesian optimization with GPs is GP-UCB \citep{srinivas10gaussian} which treats Bayesian optimization as a multi-armed bandit problem and proposes an upper-confidence bound based algorithm for query selections. The authors provide a theoretical bound on the cumulative regret that is connected with the amount of mutual information gained through the queries.
\citep{contal2014gaussian} directly incorporates mutual information into the UCB framework and demonstrated the empirical values of their method. 

Entropy search \citep{hennig2012entropy} represents another class of GP-based Bayesian optimization approach. Its main idea is to directly search for the global optimum of an unknown function through queries. Each query point is selected based on its informativeness in learning the location for the function optimum. Predictive entropy search \citep{hernandez2014predictive} addresses some computational issues from entropy search by maximizing the expected information gain with respect to the location of the global optimum. Max-value entropy search \citep{wang2016optimization,wang2017max} approaches the task of searching the global optimum differently. Instead of searching for the location of the global optimum, it looks for the value of the global optimum. This effectively avoids issues related to the dimension of the search space and the authors are able to provide regret bound analysis that the previous two entropy search methods lack.


\subsubsection{Multi-fidelitiy Bayesian Optimization}

\begin{figure}
  \centering
  \begin{subfigure}[b]{.49\textwidth}
    \centering
    {
      \includegraphics[trim={0pt 0pt 0pt 0pt}, width=.8\textwidth]{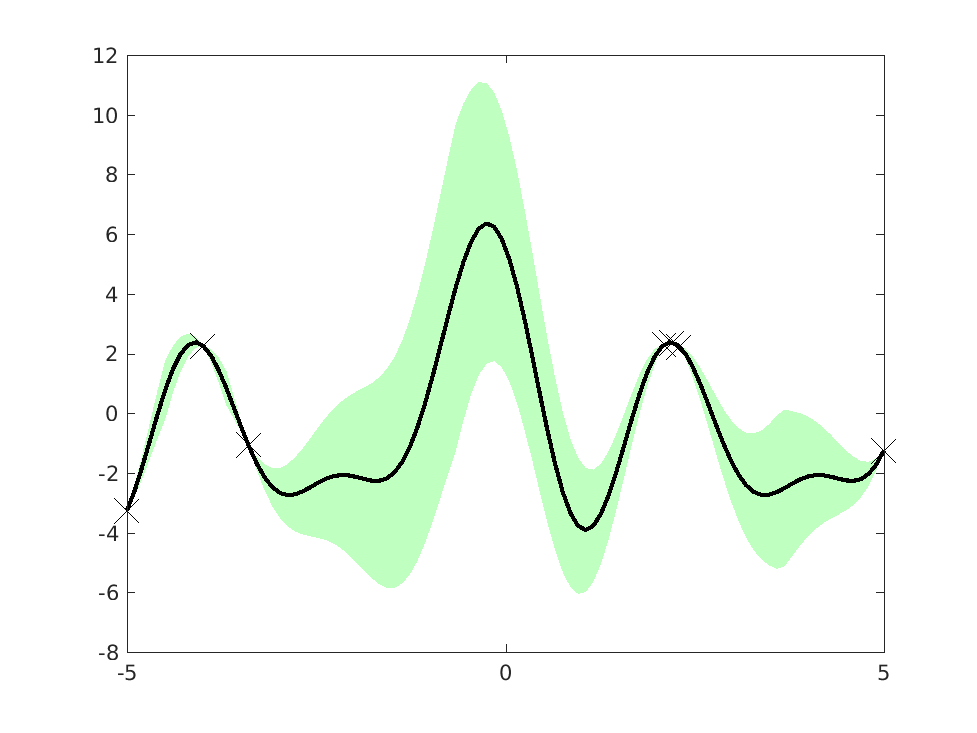}
      \caption{Only querying target fidelity function.}
      \label{fig:intro:mf:step1}
    }
  \end{subfigure}
  \begin{subfigure}[b]{.49\textwidth}
    \centering
    {
      \includegraphics[trim={0pt 0pt 0pt 0pt}, width=.8\textwidth]{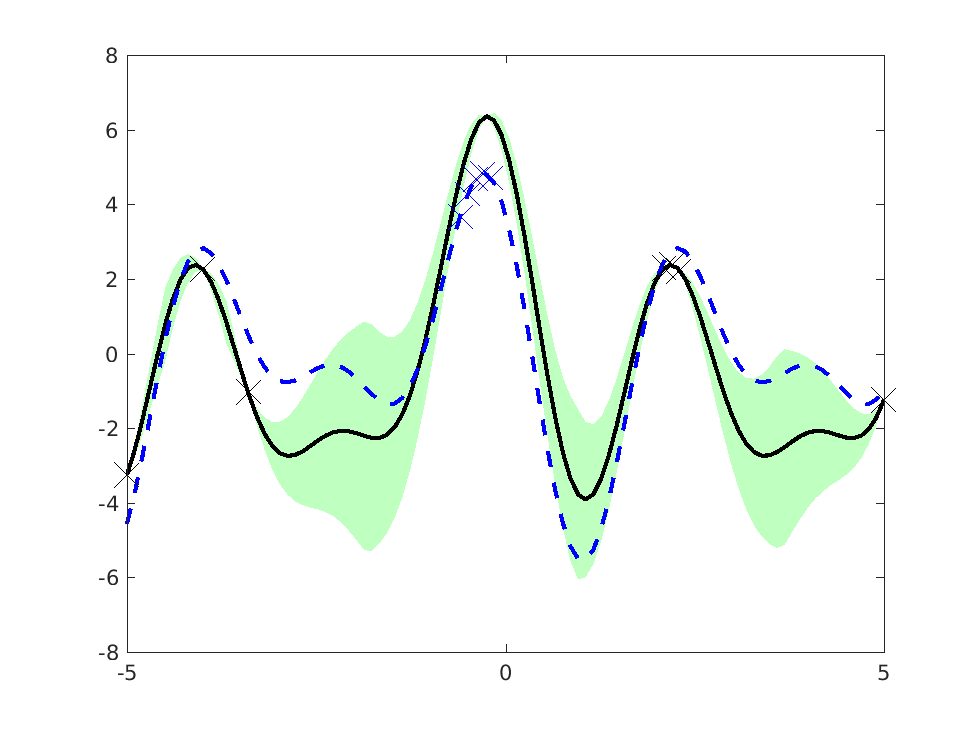}
      \caption{Querying both target and a lower fidelity.}
      \label{fig:intro:mf:step2}
    }
  \end{subfigure}
  \caption{Benefit from multi-fidelity Bayesian optimization. The left panel shows normal single fidelity Bayesian optimization where locations near a query point (crosses) have low uncertainty. When there is a lower fidelity cheaper approximation in the right panel, by querying a large number of points of the lower fidelity function, the uncertainty in the target fidelity can also be reduced significantly.}
  \label{fig:intro:multifidelity}
\end{figure}




Multi-fidelity optimization is a general framework that captures the trade-off between cheap low-quality and expensive high-quality data (cf. \figref{fig:intro:multifidelity}). There have been several works on using GPs to model functions of different fidelity levels. Recursive co-kriging \cite{forrester2007multi,le2014recursive} consider an autoregressive model for multi-fidelity GP regression, which assumes that the higher fidelity consists of a lower fidelity term and an \emph{independent} GP term which models the systematic error for approximating the higher-fidelity output. Therefore, one can model cross-covariance between the high-fidelity and low-fidelity functions using the covariance of the lower fidelity function only. \emph{Virtual vs Real} \citep{marco2017virtual} extends this idea to Bayesian optimization. The authors consider a two-fidelity setting (i.e., virtual simulation and real system experiments), where they model the correlation between the two fidelities through co-kriging, and then apply entropy search to optimize the target output.
Zhang et al. (2017) \cite{zhang2017mfpes} model the dependencies between different fidelities with convolved Gaussian processes \citep{alvarez2009sparse}, and then apply predictive entropy search (PES) \citep{hernandez2014predictive} to efficient exploration. 

Although these multi-fidelity heuristics have shown promising empirical results on some experimental datasets, little is known about their theoretical performance. Recently, Kandasamy et al. (2016) propose MF-GP-UCB (Multi-fidelity GP-UCB) \cite{kandasamy2016gaussian}, a principled for multi-fidelity Bayesian optimization. In particular, the authors consider an iterative two-stage optimization procedure and view each fidelity as an independent component, and at each iteration update the estimate of each fidelity \emph{only} based on observations from the corresponding fidelity. In a follow-up work \citep{kandasamy2017multi}, the authors address the disconnect issue by considering a continuous fidelity space and performing joint updates to effectively share information among different fidelity levels. 
However, as elaborated on in \cite{song2018general}, these approaches are likely to pick sub-optimal actions in some pessimistic cases, due to the modeling assumption and the two-stage query selection criteria. In this paper, we focus on \algname, a principled multi-fidelity algorithm as recently proposed in \cite{song2018general}. We describe the details of the algorithm in \secref{sec:method}, and evaluate it against the MF-GP-UCB algorithm as well as other single-fidelity baselines in \secref{sec:exp}.






\section{Multi-fidelity Optimization for Photonic Nanostructure Design} \label{sec:method}

\subsection{Preliminary and Problem Formulation}


Consider the problem of maximizing an unknown payoff function $\utility_\targetfid: \exDom \rightarrow \reals$.
We can probe the function $\utility_\targetfid$ by directly querying it at some $\ex \in \exDom$ and obtaining a noisy observation $\obs_{\action{\ex, \targetfid}}=\utility_\targetfid(\ex) + \noise(x)$, where $\noise(x) \sim \normal(0, \std^2)$ denotes i.i.d. Gaussian white noise. 
In addition to the payoff function $\utility_\targetfid$, we are also given access to oracle calls to some unknown auxiliary functions $\utility_1, \dots, \utility_{\targetfid-1}: \exDom \rightarrow \reals$; similarly, we obtain a noisy observation $\obs_{\action{\ex, \fid}}=\utility_\fid(\ex) + \noise$ when querying $\utility_\fid$ at $\ex$. Here, each $\utility_\fid$ could be viewed as a low-fidelity version of $\utility_\targetfid$ for $\ell<\targetfid$. For example, if $\utility_\targetfid(\ex)$ represents the actual reward obtained by running a real physical system with input $\ex$, then $\utility_\ell(\ex)$ may represent the simulated payoff from a numerical simulator at fidelity level $\fid$.


We assume that multiple fidelities $\{\utility_\fid\}_{\fid\in[m]}$ 
are mutually dependent through some fixed, (possibly) unknown joint probability distribution $\mathbb{P}[\utility_1, \dots, \utility_\targetfid]$. 
In particular,
we model $\mathbb{P}$ with a multiple output Gaussian process; hence the marginal distribution on each fidelity is a separate GP, i.e., $\forall \fid\in[\targetfid],\ \utility_\fid \sim \GP{\mean_\fid(\ex), \cov_\fid(\ex, \ex')}$, where $\mean_\fid, \cov_\fid$ specify the (prior) mean and covariance at fidelity level $\fid$. 

Let us use $\action{\ex, \fid}$ to denote the action of querying $\utility_\fid$ at $\ex$. Each action $\action{\ex, \fid}$ incurs cost $\costof{\fid}$, and achieves reward

\begin{align}
  \label{eq:reward}
  \reward(\action{\ex, \fid}) =
  \begin{cases}
    \utility_\targetfid(\ex) & \text{if}~\fid = \targetfid\\
    \reward_{\min} & \text{o.w.}
  \end{cases}
\end{align}
That is, performing $\action{\ex, \targetfid}$ 
(at the target fidelity) achieves a reward $\utility_\targetfid(\ex)$. We receive the minimal immediate reward $\reward_{\min}$ with lower fidelity actions $\action{\ex, \fid}$ for $\fid < \targetfid$, even though it may provide some information about $\utility_\targetfid$ and could thus lead to more informed decisions in the future. W.l.o.g., we assume that $\max_{\ex} \tarf(\ex) \geq 0$, and $\reward_{\min} \equiv 0$.



Let us encode an adaptive strategy for picking actions as a policy $\policy$. In words, a policy specifies which action to perform next, based on the actions picked so far and the corresponding observations. We consider policies with a fixed budget $\budget$. Upon termination, $\policy$ returns a sequence of actions $\selected_\policy$, such that $\sum_{\action{\ex, \fid} \in \selected{\policy}} \costof{\fid} \leq \budget$. Note that for a given policy $\policy$, the sequence $\selected_\policy$ is a random variable, dependent on the joint distribution $\mathbb{P}$ and the (random) observations of the selected actions.
Given a budget $\budget$ on $\policy$, our goal is to maximize the expected cumulative reward, so as to identify an action $\action{\ex, \targetfid}$ with performance close to $\ex^*=\max_{\ex \in \exDom} \utility_\targetfid(\ex)$ as rapidly as possible. Formally, we seek

\begin{align}
  \label{eq:maxreward}
  \policy^* = \argmax_{\policy: \sum_{\action{\ex, \fid} \in \selected{\policy}} \costof{\fid} \leq \budget} \expctover{\selected_\policy}{\sum_{\action{\ex, \fid} \in \selected_\policy} \reward(\action{\ex, \fid})}
\end{align}


\subsection{The \algname Algorithm}
We briefly describe \algname proposed in \citep{song2018general}, a mutual information based multi-fidelity Gaussian process optimization algorithm. It consists of two components: an exploratory procedure to gather information about the target level fidelity function via querying lower fidelity functions; an exploitative procedure to optimize the target level fidelity with the previously gathered information.
\paragraph{Exploration}
\looseness -1 \algname considers an information-theoretic selection criterion for choosing low fidelity queries. The quality of a low fidelity query $\action{\ex, \fid}$ is measured as the {\it information gain per unit cost}, defined as the amount of entropy reduction in the posterior distribution of the target fidelity function divided by the cost of the query: $\frac{\condinfgain{\obs_{\action{\ex, \fid}} }{\bobs_\selected}}{\costof{\fid}} = \frac{ \condentropy{\obs_{\action{\ex, \fid}} }{\bobs_\selected} - \condentropy{\obs_{\action{\ex, \fid}} }{\tarf, \bobs_\selected}}{\costof{\fid}}$. Here, $\selected$ denotes the set of previously selected actions, and $\bobs_{\selected}$ denote the observation history. As shown in \algref{alg:explorelf}, this criterion is used greedily to select queries for low fidelity functions. 
To ensure that the algorithm does not  explore excessively, we consider the following stopping conditions: (i) when the budget is exhausted (\lnref{alg:explorelf:ln:budget}), (ii) when a single target fidelity action is better than all the low fidelity actions in terms of the benefit-cost ratio (\lnref{alg:explorelf:ln:targetcheck}), and (iii) when the cumulative benefit-cost ratio is small (\lnref{alg:explorelf:ln:infgaincheck}). Here, the parameter $\beta$ is set to be  $\bigOmega{\frac{1}{\sqrt{B}}}$ where $B$ is the allocated budget.

\paragraph{Exploitation}
At the end of the exploration phase, \algname updates the posterior distribution of the joint GP using the full observation history, and searches for a target fidelity action via the (single-fidelity) GP optimization subroutine \sfgpopt (\lnref{ln:alg:main:sfgpout}). 
Here, \sfgpopt could be \emph{any} off-the-shelf Bayesian optimization algorithm, such as \sfgpucb \citep{srinivas10gaussian}, \sfgpmi \citep{contal2014gaussian}, \sfest \citep{wang2016optimization} and \sfmves \citep{wang2017max}, etc. Different from the previous exploration phase which seeks an informative set of low fidelity actions, the GP optimization subroutine aims to trade off exploration and exploitation on the target fidelity, and outputs a single action at each round. %
\algname then proceeds to the next round until it exhausts the preset budget, and eventually outputs an estimator of the target function optimizer.

\begin{algorithm}[t]
  \nl {\bf Input}: {Total budget $\budget$; cost $\costof{i}$ for all fidelities $i \in [\targetfid]$}; joint GP (prior) distribution on $\{\fidelity_i, \noise_i\}_{i\in[\targetfid]}$\\ 
  \Begin{
    \nl $\selected \leftarrow \emptyset$ \\
    \nl $B \leftarrow \budget$ \tcc*{initialize remainig budget}
    \While{$B > 0$} 
    {
      \tcc{explore with low fidelity}
      \nl $\eplow \leftarrow$ \explorelf $\paren{B, [\costof{\fid}], \GP{\{\utility_\fid, \noise_\fid\}_{\fid\in[\targetfid]}}, \selected}$ \\
      \tcc{select target fidelity}
      \nl $\ex^* \leftarrow \sfgpopt(\GP{\{\tarf, \noise_\targetfid\}}, \bobs_{\selected \cup \eplow})$ \label{ln:alg:main:sfgpout}\\
      \nl $\selected \leftarrow \selected \cup \eplow \cup \{\action{\ex^*,\targetfid}\}$\\
      \nl $B \leftarrow \budget - \Cost_\selected$ \tcc*{update remaining budget}
    }
    \nl {\bf Output}: Optimizer of the target function $\tarf$ \\
  }
  \caption{Multi-fidelity Mutual Information Greedy Optimization (\algname)}\label{alg:main}
\end{algorithm}

\begin{algorithm}[t]
  \nl {\bf Input}: {Exploration budget $B$; cost $[\costof{\fid}]_{\fid \in [\targetfid]}$; joint GP distribution on $\{\fidelity_i, \noise_i\}_{i\in[\targetfid]}$; previously selected items $\selected$}\\ 
  \Begin{
    \nl $\eplow \leftarrow \emptyset$ \tcc*{selected actions}
    \nl $\Cost_\eplow \leftarrow 0$ \tcc*{cost of selected actions}
    \nl $\beta \leftarrow \frac{1}{\alpha(B)}$ \tcc*{threshold}
    \While{true}
    {
      \tcc{greedy benefit-cost ratio}
      \nl $\action{\ex^*, \fid^*} \leftarrow
      \argmax_{\action{\ex, \fid}: \costof{\fid} \leq B-\Cost_\eplow - \costof{\targetfid}} \frac{\condinfgain{\obs_{\action{\ex, \fid}}}{\bobs_{\selected \cup \eplow}}}{\costof{\fid}}$ \\
      \If{$\fid^* = \nan$}
      {
        \nl break \label{alg:explorelf:ln:budget}
        \tcc*{budget exhausted}
      }
      \If{$\fid^* = \targetfid$}
      {
        \nl break \label{alg:explorelf:ln:targetcheck}
        \tcc*{worse than target}
      }
      \ElseIf{ $\frac{\condinfgain{\bobs_{\eplow \cup \{ \action{\ex^*, \fid^*} \} } }{\bobs_\selected}} {\paren{\Cost_\eplow + \costof{\fid^*}}} < \beta$}
      {
        \nl \label{alg:explorelf:ln:infgaincheck} break \tcc*{low cumulative ratio}
      }
      \Else
      {
        \nl $\eplow \leftarrow \eplow \cup \{\action{\ex^*,\fid^*}\}$ \\
        \nl $\Cost_\eplow \leftarrow \Cost_\eplow + \costof{\fid^*}$
      }
    }
  }
  \nl {\bf Output}: Selected set of items $\eplow$ from lower fidelities \\
  \caption{\explorelf: Explore low fidelities}\label{alg:explorelf}
\end{algorithm}

\subsection{Practical Implementation}
In \algref{alg:explorelf} and the algorithm used for \sfgpopt, we need to find the argmax of a function. For the photonic nanostructure experiment in \secref{sec:exp}, this optimization is over a discrete set of candidate queries. Naively, we would need to evaluate the function at each query point in order to determine the optimizer, which is a costly operation. Instead, we devise an approximate optimization step to address this computational challenge. We first directly optimize the function over its {\it continuous} domain and obtain an optimizer. Then we project the optimizer down to the candidate set by selecting the closest available query point based on Euclidean distance. This approximation scheme takes advantage of existing fast optimizers for continuous functions and becomes necessary for large candidate size.


\section{Experimental Design for Photonic Nanostructures Discovery}
\label{sec:exp}

\subsection{Datasets}

Our nanophotonic structure is characterized by the five geometric parameters. For each parameter setting, we use a score, commonly called a figure-of-merit (FOM), to represent how well the resulting structure satisfies the desired color filtering property. By minimizing FOM, we can find a set of high-quality design parameters. Traditionally, FOMs can only be computed through the actual fabrication of a structure and subsequent measurements of its various physical properties, which is a time-consuming process. Alternatively, simulations can be utilized to estimate what physical properties a design will have, e.g. using the Lumerical software. By solving a 2D variant of Maxwell's equations, we could simulate the transmission spectrum of a given nanophotonic device and then compute FOM from it. We could obtain different fidelity level data by controlling aspects of the numerical solution process.

We experiment with three design tasks for filtering light with wavelengths of 550 nm, 650 nm and 750 nm. For each task, we vary the conformal mesh size and the time-domain solver total time duration of simulated physical processes to obtain two sets of multi-fidelity data, each with three fidelity levels on 4983 designs.

The first set of data is based on different conformal mesh sizes. The mesh size determines how accurate the final results are, with finer meshes lead to more accurate results. We generated the lowest fidelity data using a mesh size of $3\text{nm}\times 3\text{nm}$, the middle fidelity $2\text{nm}\times 2\text{nm}$ and the target fidelity $1\text{nm}\times 1\text{nm}$. The costs, CPU time, are inverse proportional to the mesh size, so we use the following costs $[1, 2.25, 9]$ for our three fidelity function evaluation, respectively.

The second set of data is based on the different total time duration of simulated physical processes for the time-domain solver. Since the transmission spectrum is calculated through Fourier transform of the electromagnetic pulse, that is passed through the color filter, we expect more accurate solutions with longer physical simulation time duration. We generated the lowest fidelity data using a simulation time of 40 fs (femtoseconds), the middle fidelity 70 fs and the target fidelity 100 fs. The costs are proportional to the simulation time, so we use the following costs $[40, 70, 100]$ for our three fidelity function evaluation, respectively.

\subsection{Experimental Setup}
\looseness -1 To model the relationship between a low fidelity function $\utility_{i}$ and the target fidelity function $\tarf$, we use an additive model. Specifically, we assume that $\utility_i = \tarf + \noise_i$ for all fidelity levels $i < m$ where $\noise_i$ is an unknown function characterizing the error incurred by a lower fidelity function. We use Gaussian processes to model $\tarf$ and $\noise_i$. Since $\tarf$ is embedded in every fidelity level, we can use an observation from any fidelity to update the posterior for \emph{every} fidelity level. We use square exponential kernels for all the GP covariances, with hyperparameter tuning scheduled periodically during optimization. Following prior work on practical Bayesian optimization \citep{brochu2010tutorial}, we use 10\% of the total budget for initialization. For multi-fidelity methods, the initialization budget is spent on randomly querying the lowest fidelity function. For the single-fidelity method, it is spent on randomly querying the target fidelity function. For all experiments, we use a total budget of 100 times the cost of target fidelity function call $\tarf$. Every method is run 20 times to compute its mean and standard error.

\subsection{Compared Methods}
Our framework is general and we could plug in different single fidelity Bayesian optimization algorithms for the \sfgpopt procedure in Algorithm \ref{alg:main}. In our experiment, we choose to use GP-UCB as one instantiation. We compare with MF-GP-UCB \citep{kandasamy2016gaussian} and GP-UCB \citep{srinivas10gaussian}. MF-GP-UCB relies on several hyperparameters in the algorithm, we keep the same approach to choosing them as described in \citep{kandasamy2016gaussian}. 

Besides the Bayesian optimization based method, we also compare with a common heuristic called Particle Swarm Optimization, which is inspired by the social behavior of animals and is used for nanophotonic structure designs \citep{pso2008mutitu,shokooh2010leaky}. We use built-in MATLAB implementation of this algorithm. We specify a population of 5 particles and run Swarm optimization for 20 iterations, totaling $5\times 20 = 100$ evaluations of the target fidelity function. All other algorithm parameters are kept at default MATLAB values.



\begin{figure*}[t]
  \centering
  \begin{subfigure}[b]{.32\textwidth}
    \centering
    {
      \includegraphics[trim={0pt 0pt 0pt 0pt}, width=\textwidth]{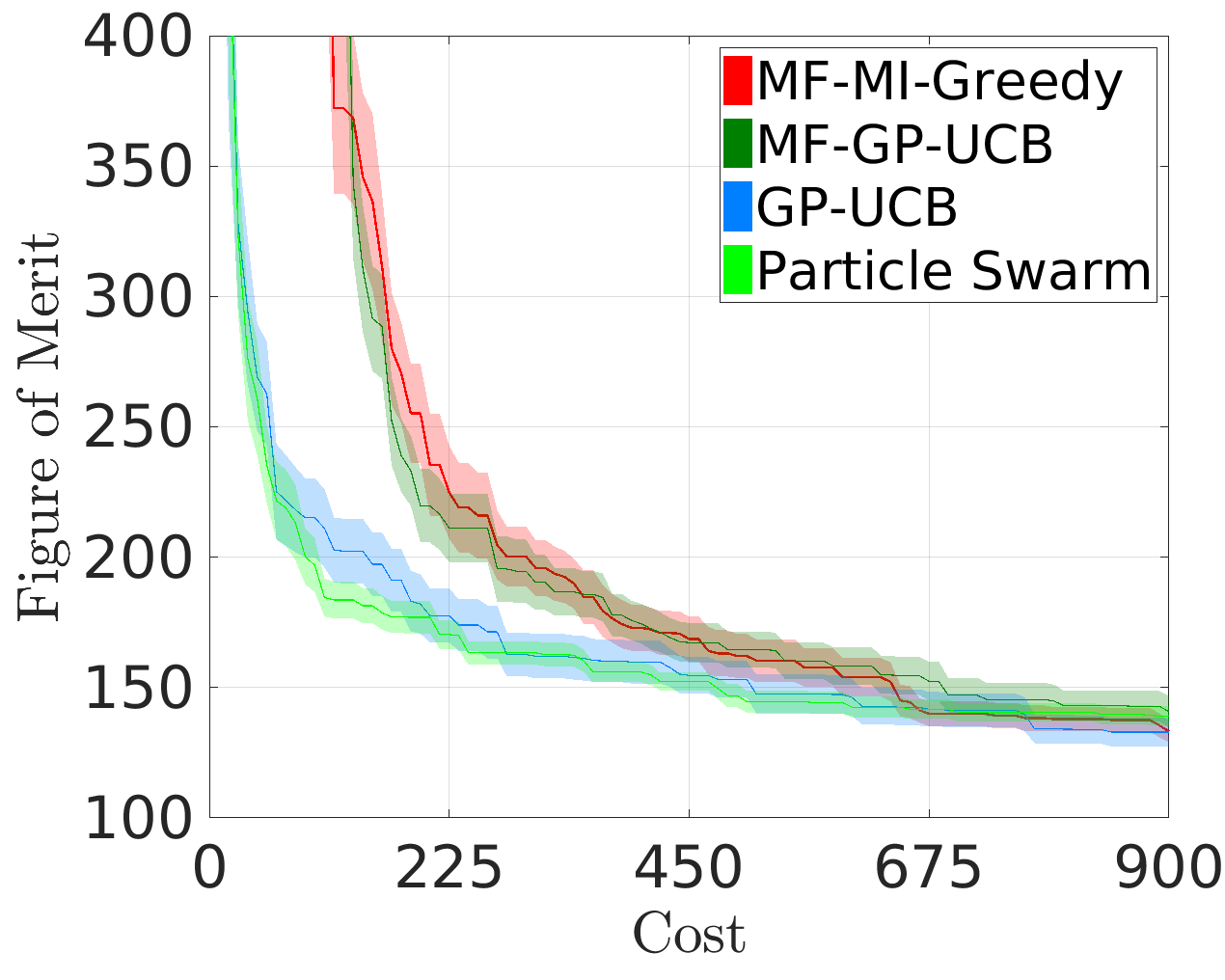}
      \caption{550nm wavelength}
      \label{fig:real:mesh:550nm}
    }
  \end{subfigure}
  \begin{subfigure}[b]{.32\textwidth}
    \centering
    {
      \includegraphics[trim={0pt 0pt 0pt 0pt}, width=\textwidth]{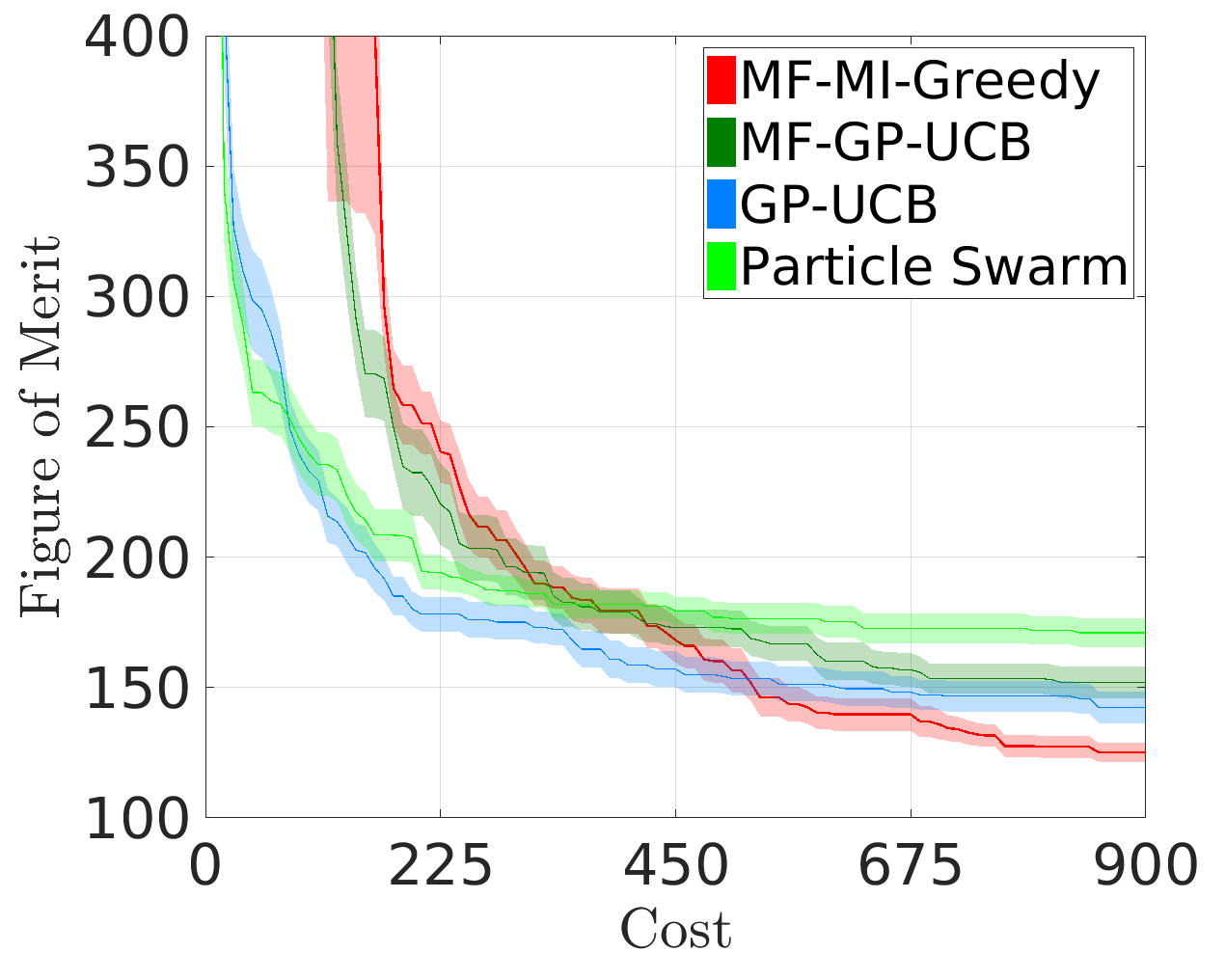}
      \caption{650nm wavelength}
      \label{fig:real:mesh:650nm}
    }
  \end{subfigure}
  \begin{subfigure}[b]{.32\textwidth}
    \centering
    {
      \includegraphics[trim={0pt 0pt 0pt 0pt}, width=\textwidth]{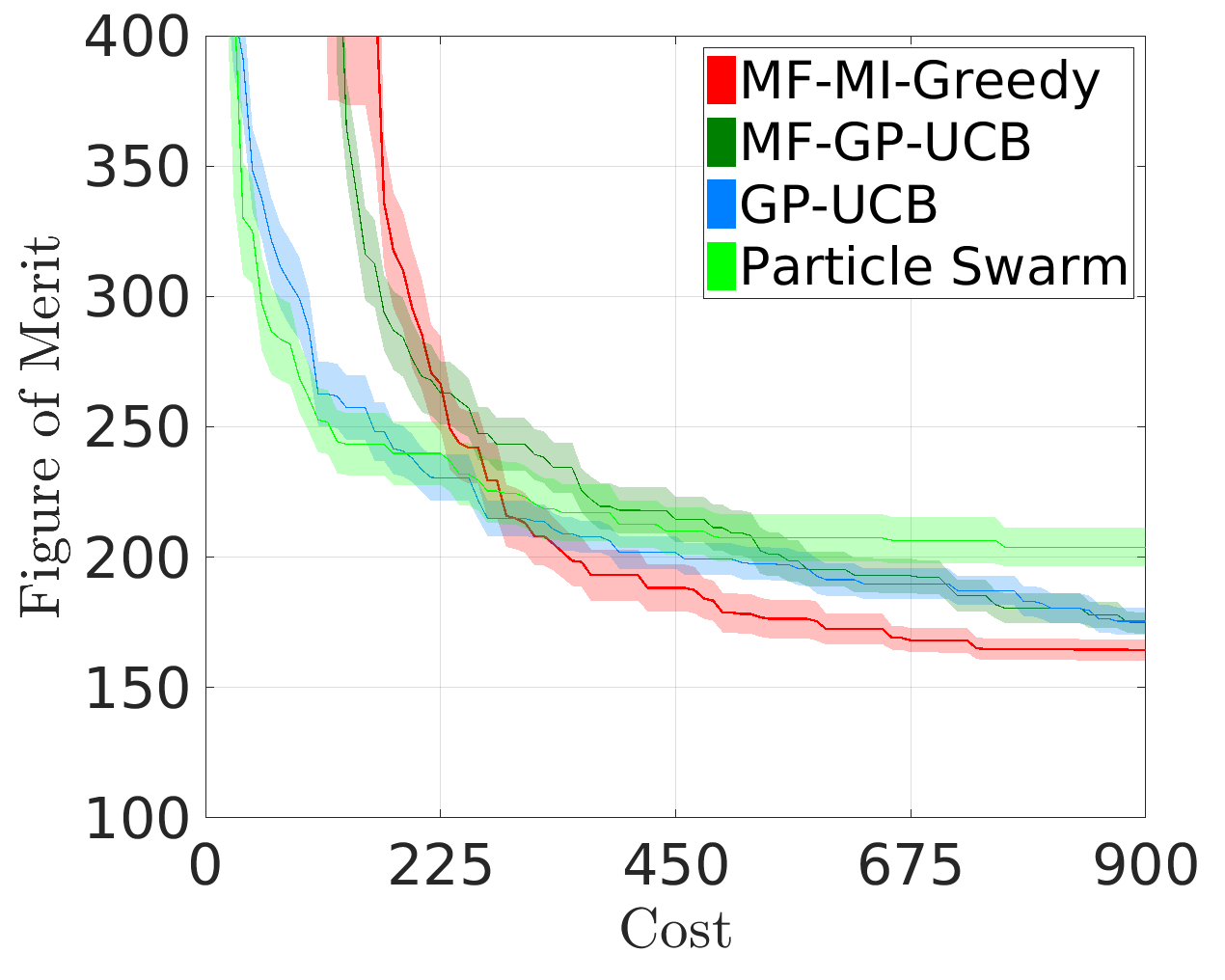}
      \caption{750nm wavelength}
      \label{fig:real:mesh:750nm}
    }
  \end{subfigure}
    \caption{Multi-fidelity based on conformal mesh size. Every method is run 20 times and we plot the mean plus/minus one standard error in the figures.}
    \label{fig:exp:real:mesh}
\end{figure*}

\begin{figure*}[t]
  \centering
  \begin{subfigure}[b]{.32\textwidth}
    \centering
    {
      \includegraphics[trim={0pt 0pt 0pt 0pt}, width=\textwidth]{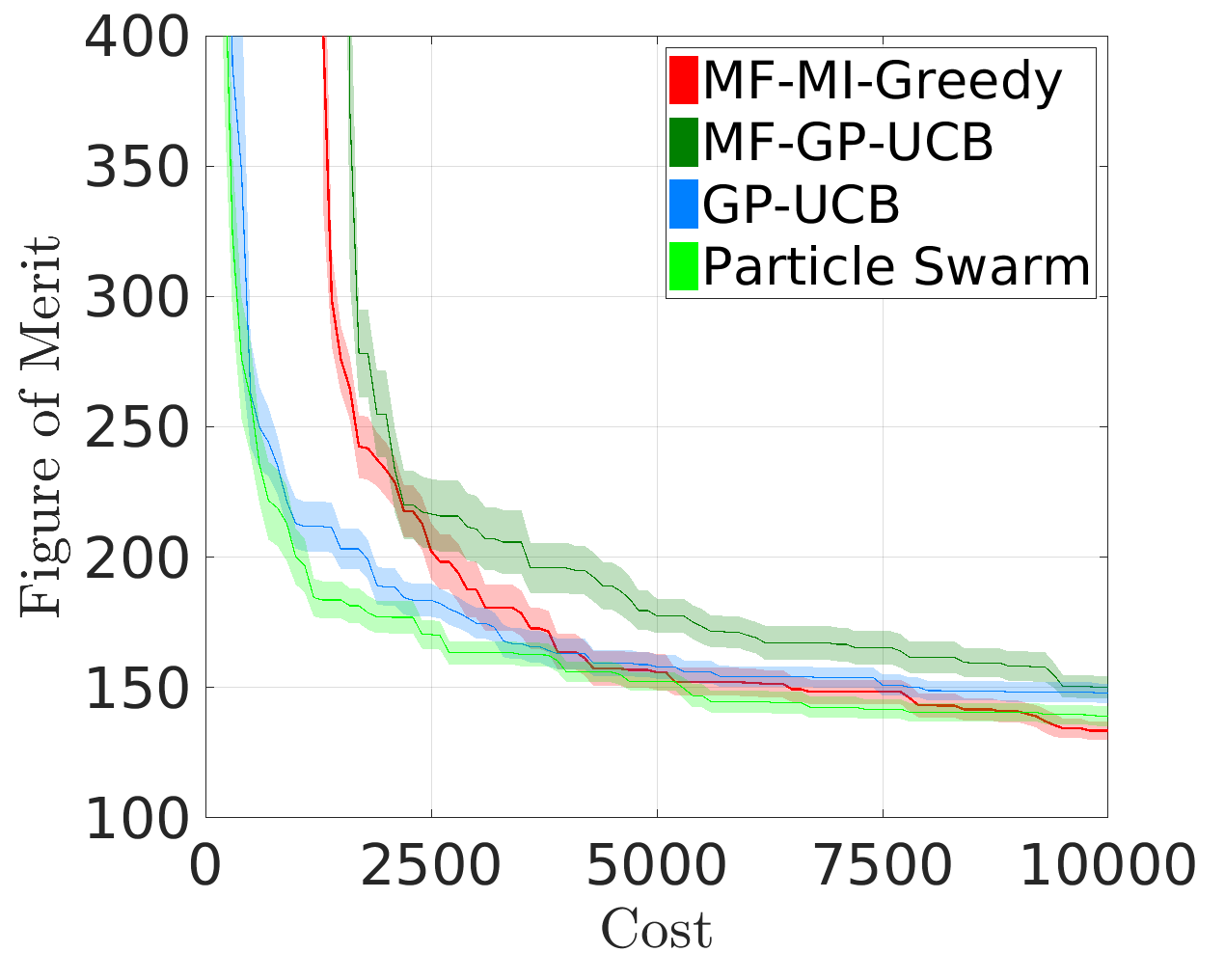}
      \caption{550nm wavelength}
      \label{fig:real:time:550nm}
    }
  \end{subfigure}
  \begin{subfigure}[b]{.32\textwidth}
    \centering
    {
      \includegraphics[trim={0pt 0pt 0pt 0pt}, width=\textwidth]{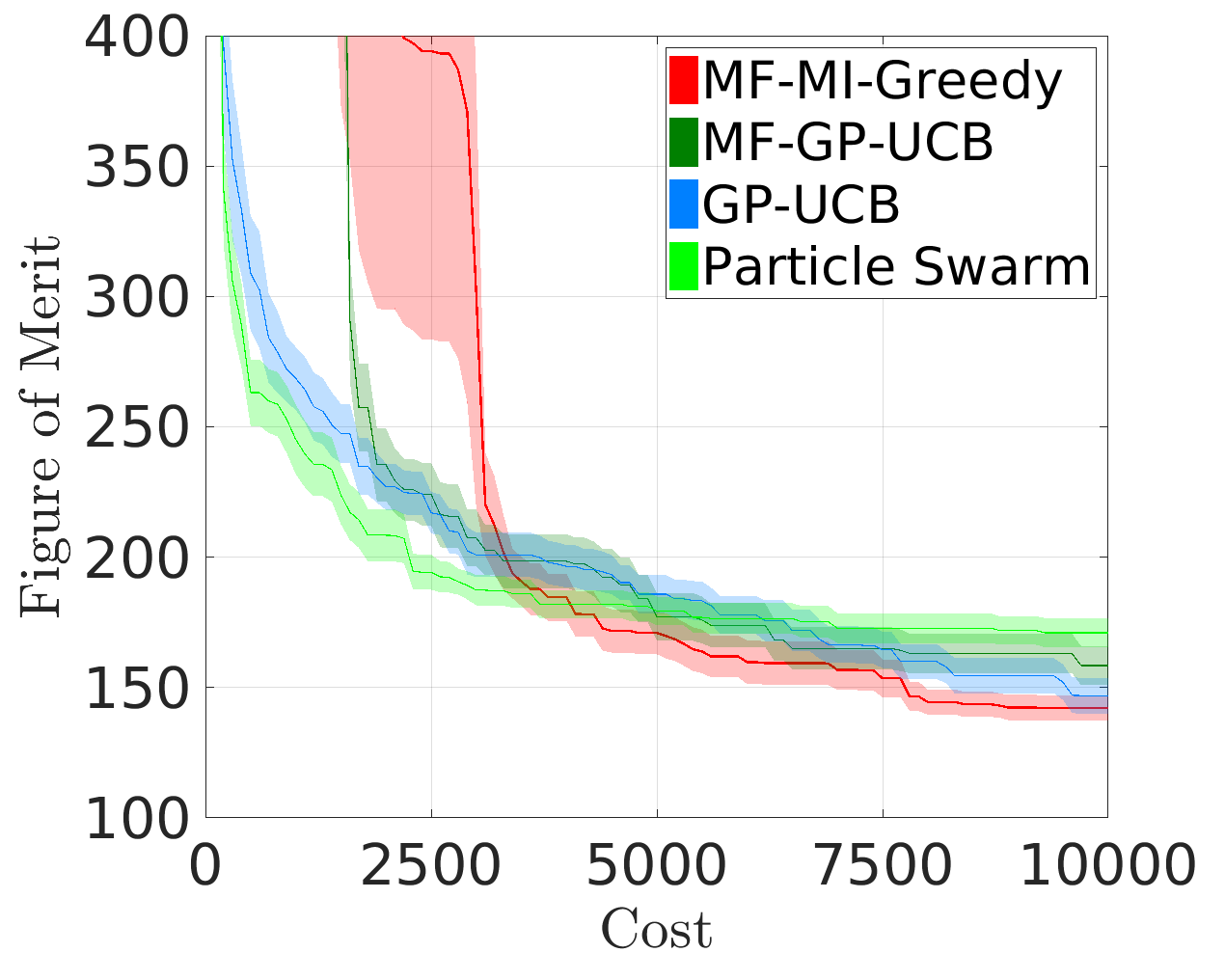}
      \caption{650nm wavelength}
      \label{fig:real:time:650nm}
    }
  \end{subfigure}
  \begin{subfigure}[b]{.32\textwidth}
    \centering
    {
      \includegraphics[trim={0pt 0pt 0pt 0pt}, width=\textwidth]{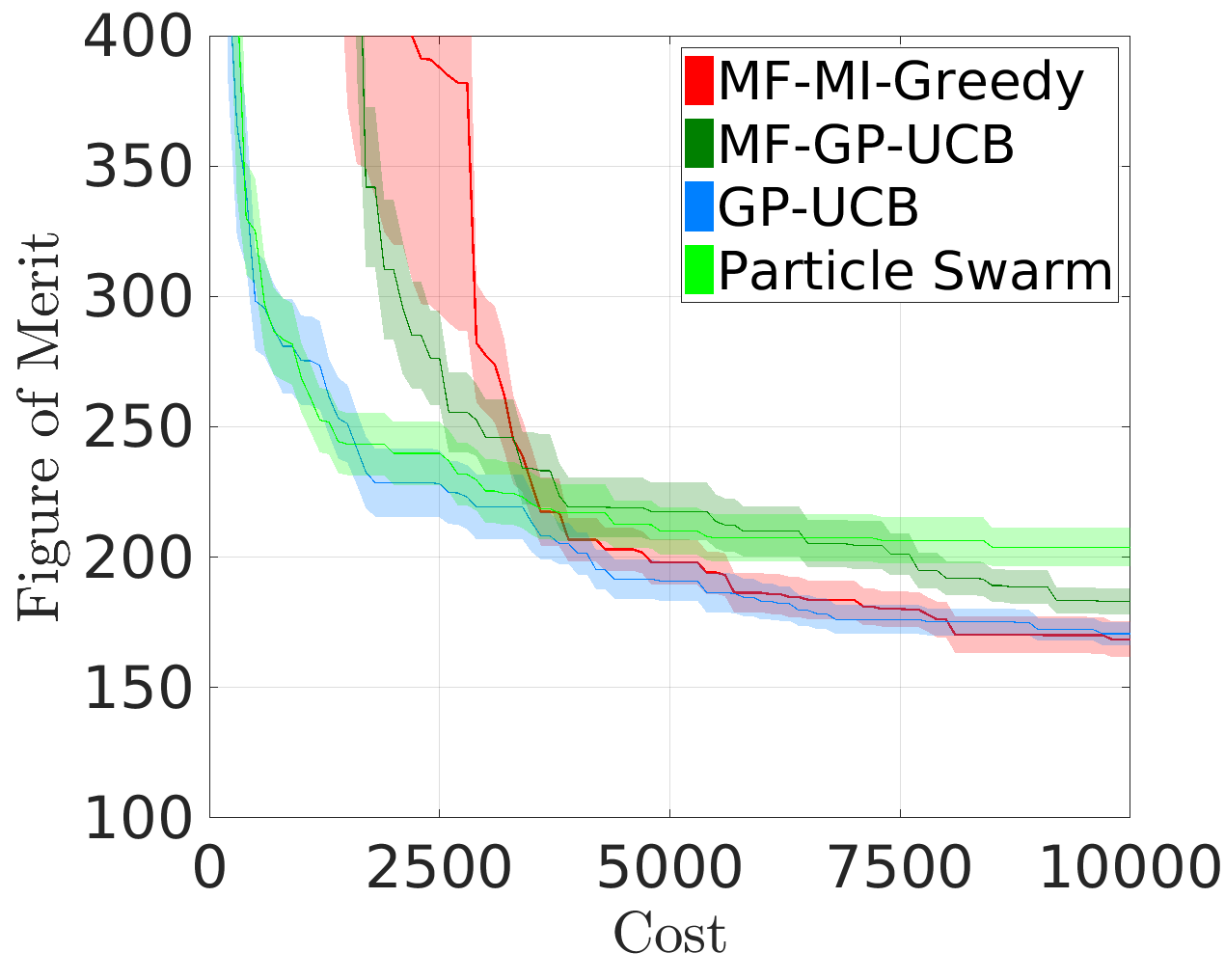}
      \caption{750nm wavelength}
      \label{fig:real:time:750nm}
    }
  \end{subfigure}
    \caption{Multi-fidelity based on conformal simulation time. Every method is run 20 times and we plot the mean plus/minus one standard error in the figures.}
    \label{fig:exp:real:time}
\end{figure*}

\subsection{Optimizing Figure of Merit}
Figure \ref{fig:exp:real:mesh} and Figure \ref{fig:exp:real:time} show the results of this experiment. As usual, the $x$-axis is the cost and $y$-axis is Figure of Merit and smaller is better. After a small portion of the budget is used in initial exploration, MF-MI-Greedy (red) is able to arrive at a better final design compared with MF-GP-UCB, GP-UCB and Particle Swarm. MF-MI-Greedy tends to have a worse figure of merit at the beginning because the initial explorations in the lower fidelity do not yield FOM scores on the target fidelity, so essentially it has a late start in all the plots because it starts querying the target fidelity late. However, the advantage of exploring lower fidelities becomes apparent once the exploitation phase starts in the target fidelity level, as seen by the rapid convergence to low FOM designs.


\section{Conclusion}
In this paper, we investigate the problem of optimizing the transmission properties of plasmonic mirror color filters, where we have access to multiple numerical simulators with different fidelity levels and computational costs. We considered several derivative-free global optimization algorithms, including a commonly used approach in the nanophotonics community, and two recently developed multiple-fidelity Bayesian optimization approaches. Our results on several pre-collected nanophotonics datasets demonstrate the compelling performance of the multiple-fidelity Bayesian optimization approach. These experiments suggest that there is a great potential in utilizing cheap, multi-fidelity simulations to aid the discovery of the optimal photonic nanostructures.



\bibliography{reference}

\begin{thebibliography}{10}

\bibitem{alvarez2009sparse}
Mauricio Alvarez and Neil~D Lawrence.
\newblock Sparse convolved gaussian processes for multi-output regression.
\newblock In {\em Advances in neural information processing systems}, pages
  57--64, 2009.

\bibitem{brochu2010tutorial}
Eric Brochu, Vlad~M Cora, and Nando De~Freitas.
\newblock A tutorial on bayesian optimization of expensive cost functions, with
  application to active user modeling and hierarchical reinforcement learning.
\newblock {\em arXiv preprint arXiv:1012.2599}, 2010.

\bibitem{chen2012refl}
Jing Chen, Jian Yang, Zhuo Chena, Yi-Jiao Fang, Peng Zhan, and Zhen-Lin Wang.
\newblock Plasmonic reflectors and high-q nano-cavities based on coupled
  metal-insulator-metal waveguides.
\newblock {\em AIP Advances}, 2:012145, 2012.

\bibitem{contal2014gaussian}
Emile Contal, Vianney Perchet, and Nicolas Vayatis.
\newblock Gaussian process optimization with mutual information.
\newblock In {\em International Conference on Machine Learning}, pages
  253--261, 2014.

\bibitem{ebbesen1998eot}
Thomas~W. Ebbesen, Henri~J. Lezec, H.~F. Ghaemi, Tineke Thio, and Peter~A.
  Wolff.
\newblock Extraordinary optical transmission through sub-wavelength hole
  arrays.
\newblock {\em Nature}, 391:667--668, 1998.

\bibitem{dagny2017filter}
Dagny Fleischman, Luke~A. Sweatlock, Hirotaka Murakami, and Harry Atwater.
\newblock Hyper-selective plasmonic color filters.
\newblock {\em Optics Express}, 25:27386--27395, 2017.

\bibitem{forrester2007multi}
Alexander I.~J. Forrester, Andr{\'a}s S{\'o}bester, and Andy~J. Keane.
\newblock Multi-fidelity optimization via surrogate modelling.
\newblock In {\em Proceedings of the royal society of london a: mathematical,
  physical and engineering sciences}, volume 463, pages 3251--3269. The Royal
  Society, 2007.

\bibitem{hennig2012entropy}
Philipp Hennig and Christian~J. Schuler.
\newblock Entropy search for information-efficient global optimization.
\newblock {\em Journal of Machine Learning Research}, 13(Jun):1809--1837, 2012.

\bibitem{hernandez2014predictive}
Jos{\'e}~Miguel Hern{\'a}ndez-Lobato, Matthew~W. Hoffman, and Zoubin
  Ghahramani.
\newblock Predictive entropy search for efficient global optimization of
  black-box functions.
\newblock In {\em Advances in neural information processing systems}, pages
  918--926, 2014.

\bibitem{holland1984genetic}
John~H. Holland.
\newblock Genetic algorithms and adaptation.
\newblock {\em Adaptive Control of Ill-Defined Systems. NATO Conference Series
  (II Systems Science)}, 16, 1984.

\bibitem{kandasamy2016gaussian}
Kirthevasan Kandasamy, Gautam Dasarathy, Junier~B. Oliva, Jeff Schneider, and
  Barnab{\'a}s P{\'o}czos.
\newblock Gaussian process bandit optimisation with multi-fidelity evaluations.
\newblock In {\em Advances in Neural Information Processing Systems}, pages
  992--1000, 2016.

\bibitem{kandasamy2017multi}
Kirthevasan Kandasamy, Gautam Dasarathy, Jeff Schneider, and Barnab{\'a}s
  P{\'o}czos.
\newblock Multi-fidelity bayesian optimization with continuous approximations.
\newblock In {\em International Conference on Machine Learning}, pages
  1799--1808, 2017.

\bibitem{kennedy1995pso}
James Kennedy and Russel~C. Eberhart.
\newblock Particle swarm optimization.
\newblock {\em Proceedings of the IEEE International Conference on Neural
  Networks. Perth, Australia}, page 1942–1945, 1995.

\bibitem{kirkpatrick1983annealing}
Scott Kirkpatrick, Charles~Daniel Gelatt, and Mario~P. Vecchi.
\newblock Optimization by simulated annealing.
\newblock {\em Science}, 220:671--680, 1983.

\bibitem{le2014recursive}
Loic Le~Gratiet and Josselin Garnier.
\newblock Recursive co-kriging model for design of computer experiments with
  multiple levels of fidelity.
\newblock {\em International Journal for Uncertainty Quantification}, 4(5),
  2014.

\bibitem{atwater2001plasmonics}
Stefan~A. Maier, Mark~L. Brongersma, Pieter~G. Kik, Sheffer Meltzer, Ari A.~G.
  Requicha, and Harry~A. Atwater.
\newblock Plasmonics---a route to nanoscale optical devices.
\newblock {\em Advanced materials}, 13:1501--1505, 2001.

\bibitem{marco2017virtual}
Alonso Marco, Felix Berkenkamp, Philipp Hennig, Angela~P. Schoellig, Andreas
  Krause, Stefan Schaal, and Sebastian Trimpe.
\newblock Virtual vs. real: Trading off simulations and physical experiments in
  reinforcement learning with bayesian optimization.
\newblock In {\em 2017 IEEE International Conference on Robotics and Automation
  (ICRA)}, pages 1557--1563, 2017.

\bibitem{pso2008mutitu}
James~G. Mutitu, Shouyuan Shi, Caihua Chen, Timothy Creazzo, Allen Barnett,
  Christiana Honsberg, and Dennis~W Prather.
\newblock Thin film silicon solar cell design based on photonic crystal and
  diffractive grating structures.
\newblock {\em Optics express}, 16(19):15238--15248, 2008.

\bibitem{rasmussen:williams:2006}
Carl~E. Rasmussen and Chris K.~I. Williams.
\newblock {\em Gaussian Processes for Machine Learning}.
\newblock MIT Press, 2006.

\bibitem{shokooh2010leaky}
Mehrdad Shokooh-Saremi and Robert Magnusson.
\newblock Leaky-mode resonant reflectors with extreme bandwidths.
\newblock {\em Optics letters}, 35(8):1121--1123, 2010.

\bibitem{snoek2012practical}
Jasper Snoek, Hugo Larochelle, and Ryan~P. Adams.
\newblock Practical bayesian optimization of machine learning algorithms.
\newblock In {\em Advances in neural information processing systems}, pages
  2951--2959, 2012.

\bibitem{song2018general}
Jialin Song, Yuxin Chen, and Yisong Yue.
\newblock A general framework for multi-fidelity bayesian optimization with
  gaussian processes.
\newblock {\em arXiv preprint arXiv:1811.00755}, 2018.

\bibitem{srinivas10gaussian}
Niranjan Srinivas, Andreas Krause, Sham Kakade, and Matthias Seeger.
\newblock Gaussian process optimization in the bandit setting: No regret and
  experimental design.
\newblock In {\em Proc. International Conference on Machine Learning (ICML)},
  2010.

\bibitem{wang2017max}
Zi~Wang and Stefanie Jegelka.
\newblock Max-value entropy search for efficient bayesian optimization.
\newblock {\em arXiv preprint arXiv:1703.01968}, 2017.

\bibitem{wang2016optimization}
Zi~Wang, Bolei Zhou, and Stefanie Jegelka.
\newblock Optimization as estimation with gaussian processes in bandit
  settings.
\newblock In {\em Artificial Intelligence and Statistics}, pages 1022--1031,
  2016.

\bibitem{yokogawa2013filter}
Sozo Yokogawa, Stanley~P. Burgos, and Harry~A. Atwater.
\newblock Plasmonic color filters for \textsc{CMOS} image sensor applications.
\newblock {\em Nano. Lett., ACS Nano}, 7:10038--10047, 2013.

\bibitem{zhang2017mfpes}
Yehong Zhang, Trong~Nghia Hoang, Bryan Kian~Hsiang Low, and Mohan Kankanhalli.
\newblock Information-based multi-fidelity bayesian optimization.
\newblock {\em NIPS Workshop on Bayesian Optimization}, 2017.

\end{thebibliography}


\end{document}